\documentclass[letterpaper, 10 pt, conference]{ieeeconf}  

\IEEEoverridecommandlockouts                              

\usepackage{graphicx}
\usepackage{epsfig} 
\usepackage{mathptmx} 
\usepackage{times} 
\usepackage{amsmath} 
\usepackage{amssymb}  
\usepackage{booktabs}
\usepackage{bm}
\usepackage{cite}

\usepackage{etoolbox}
\makeatletter
\patchcmd{\@makecaption}
  {\scshape}
  {}
  {}
  {}
\makeatletter
\patchcmd{\@makecaption}
  {\\}
  {:\ }
  {}
  {}
\makeatother

\newcommand{\Chk}{{\small$\checkmark$}}
\newcommand{\Mns}{{\small$-$}}
\newcommand{\ChkMini}{{\tiny$\checkmark$}}
\newcommand{\MnsMini}{{\tiny$-$}}

\usepackage{url}
\usepackage{xspace}

\title{\LARGE \bf
LiTAMIN2: Ultra Light LiDAR-based SLAM using \\Geometric Approximation applied with KL-Divergence
}

\author{Masashi Yokozuka$^{1}$, Kenji Koide$^{1}$, Shuji Oishi$^{1}$ and Atsuhiko Banno$^{1}$
\thanks{$^{1}$The authors are with the Human-Centered Mobility Research Center (HCMRC),
        National Institute of Advanced Industrial Science and Technology (AIST), Japan
        {\tt\small yokotsuka-masashi@aist.go.jp}}%
\thanks{This work was supported in part by the New Energy and Industrial Development Organization (NEDO).}%
}

\begin{document}
\bstctlcite{IEEEexample:BSTcontrol}

\maketitle
\thispagestyle{empty}
\pagestyle{empty}

\begin{abstract}

In this paper, a three-dimensional  light detection and ranging simultaneous localization and mapping (SLAM) method is proposed that is available for tracking and mapping with 500--1000 Hz processing. 
The proposed method significantly reduces the number of points used for point cloud registration using a novel ICP metric to speed up the registration process while maintaining accuracy. 
Point cloud registration with ICP is less accurate when the number of points is reduced because ICP basically minimizes the distance between points. 
To avoid this problem, symmetric KL-divergence is introduced to the ICP cost that reflects the difference between two probabilistic distributions. 
The cost includes not only the distance between points but also differences between distribution shapes. 
The experimental results on the KITTI dataset indicate that the proposed method has high computational efficiency, strongly outperforms other methods, and has similar accuracy to the state-of-the-art SLAM method. 

\end{abstract}


%
{
\begin{figure}[!t]
\centering
\includegraphics[width=1.0\linewidth]{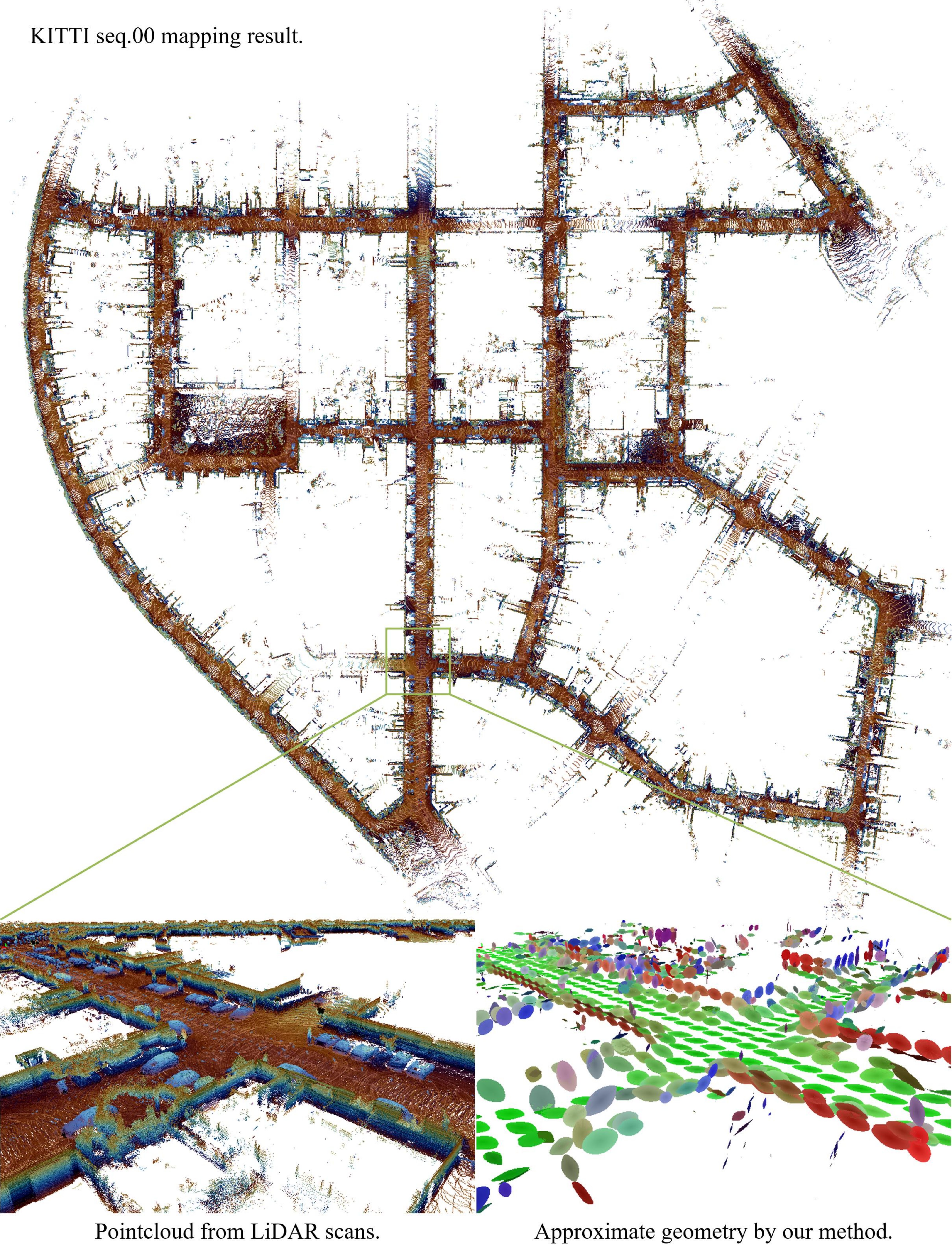}
\caption{Example mapping result for the KITTI sequence 00 data using LiTAMIN2. The color of the bottom right figure 
indicates the normal direction given from normal distributions decomposed using principal component analysis, i.e. the direction is the eigenvector of the minimum eigenvalue. }
\label{fig:Main}
\vspace{-2mm}
\end{figure}
}

\section{Introduction}
Simultaneous localization and mapping (SLAM) is a fundamental element of mobility technologies and services, such as autonomous mobile robots.
In particular, light detection and ranging (LiDAR) and depth sensors have already been commercialized and are being applied because of their stable and accurate performance.
In the near future, not only self-driving cars, but all types of mobile devices will be equipped with LiDAR or depth sensors.
We anticipate a world in which point cloud data captured via SLAM will be aggregated in the cloud and shared to provide a variety of services.

There is a need to efficiently generate and update global maps from the huge amount of point cloud data aggregated in real time from devices around the world.
Because the number of servers used in this process is much smaller than the number of devices, it is essential to use SLAM methods that go beyond real-time performance.
The performance of current LiDAR-based SLAM is only slightly better than real-time performance.

In addition to the server, speedup is also necessary to operate SLAM on edge devices, which are severely constrained in terms of computational resources.
The current LiDAR-based SLAM method is based on the premise that real-time performance is guaranteed using the CPU and GPU on a PC, and it is necessary to improve the computational efficiency of the SLAM method to ensure real-time performance on edge devices.

Although many studies have been conducted on SLAM benchmarks \cite{KITTI} and methods that emphasize accuracy \cite{LiTAMIN,hdl-graph-SLAM,SuMa-SLAM,Elastic-SLAM,ElasticFusion,VelodyneSLAM,ContSLAM,LOAM-SLAM,LOAM-SLAM-Paper,LeGO-LOAM-SLAM,LIO-SLAM,LIO-SAM,LINS}, few studies have significantly improved the current computational efficiency.
In the future, to process a large number of robots and devices intensively and efficiently, it is expected that SLAM will emphasize high speed.

The aim of this paper is to establish a method that is as accurate as the state-of-the-art method, while significantly exceeding real-time performance.
In this paper, a three-dimensional  (3D) LiDAR-based SLAM is discussed, which significantly improves the computational efficiency of LiDAR-based SLAMs, running at 500--1000 Hz and providing the same level of accuracy as state-of-the-art methods.
The proposed method significantly reduces the number of points used for point cloud registration using a novel ICP metric while maintaining accuracy. 
Point cloud registration with ICP is less accurate when the number of points is reduced. 
To avoid this problem, symmetric KL-divergence is introduced to the ICP cost. 
The experimental results (Fig. \ref{fig:Main}) on the KITTI dataset indicate that the proposed method has high computational efficiency, strongly outperforms other methods, and has similar accuracy to the state-of-the-art SLAM method. 

{
\begin{figure*}[!t]
\centering
\vspace{2mm}
\includegraphics[width=1.0\linewidth]{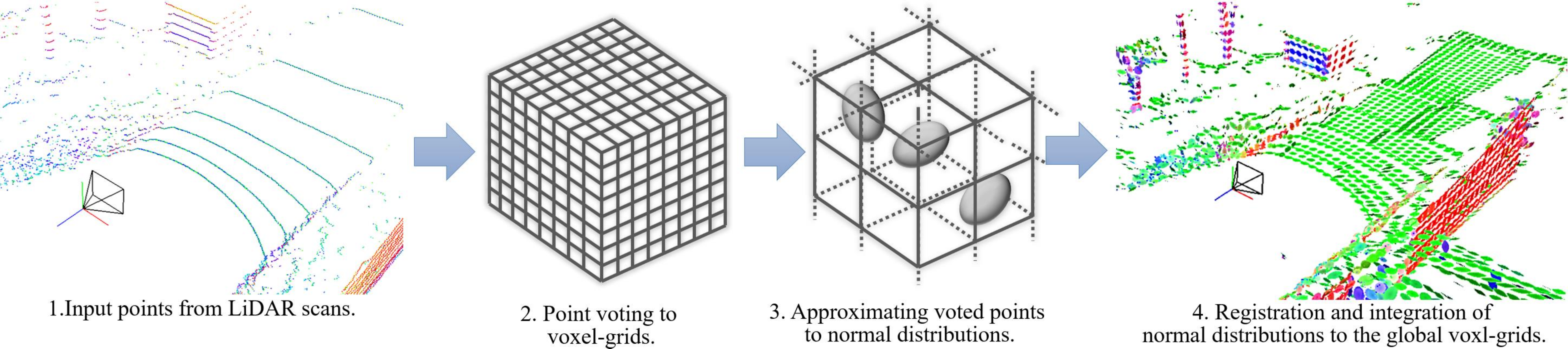}
\caption{Overview of our method. The color of the figure on the right shows the normal direction given from normal distributions decomposed using principal component analysis; the color is similar to that in Figure \ref{fig:Main}.}
\label{fig:overview}
\vspace{-1mm}
\end{figure*}
}

{
\begin{table*}[!t]\renewcommand{\arraystretch}{1.6}
\begin{center}
\caption{Comparison of the ICP cost functions for local approximation with a cluster of normal distributions.}
\label{tbl:ICP}
\setlength{\tabcolsep}{3.0mm}
{\footnotesize
\begin{tabular}{c|c}
\toprule
Method & ICP cost function per point-assosiation    \\ \midrule[1pt]
Standard ICP                & $\left(q- (R p + t)\right)^T \left(q- (R p + t)\right)$ \\ \hline
NDT                         & $\left(q- (R p + t)\right)^T C^{-1} \left(q- (R p + t)\right)$ \\ \hline
Generalized ICP             & $\left(q- (R p + t)\right)^T (C_{q}+R C_{p} R^T)^{-1} \left(q- (R p + t)\right)$ \\ \hline
LiTAMIN                     & $\left(q- (R p + t)\right)^T \cfrac{w (C + \lambda I)^{-1}}{\|(C + \lambda I)^{-1}\|_F} \left(q- (R p + t)\right)$ \\ \hline
LiTAMIN2 (proposed method)  & $
%
			w_{ICP} \left[ \left(q- (R p + t)\right)^T \cfrac{(C_{q}+R C_{p} R^T + \lambda I)^{-1}}{\|(C_{q}+R C_{p} R^T + \lambda I)^{-1}\|_F} \left(q- (R p + t)\right) \right] + 
			w_{Cov} \left[ \mathrm{Tr}(R C_{p}^{-1} R^T C_{q}) + \mathrm{Tr}(C_{q}^{-1} R C_{p} R^T) -6 \right] $\\ 
%
\bottomrule
\end{tabular}
}
\end{center}
\vspace{-1mm}
\end{table*}
}

\section{Related work}

LiDAR SLAM methods can be divided into two categories: ICP-based methods\cite{LiTAMIN,hdl-graph-SLAM,SuMa-SLAM,Elastic-SLAM,ElasticFusion,VelodyneSLAM,ContSLAM} and feature-based methods\cite{LOAM-SLAM,LOAM-SLAM-Paper,LeGO-LOAM-SLAM,LIO-SLAM,LIO-SAM,LINS}.

For ICP-based methods, voxelization is a simple but effective approach to speed it up.
By dividing the point clouds into small groups and approximating each sub-point cloud with a normal distribution, the number of points can be significantly reduced while preserving the shape, to a certain extent.
Normal distribution transformation (NDT)\cite{NDT-ICP,Takeuchi-ICP,NDTvs-ICP} and generalized-ICP (GICP)\cite{G-ICP} are the most common ICP methods for performing voxel approximation, but there are some differences between them.
NDT approximates only the target points with normal distributions and determines the voxel-wise correspondences,
whereas GICP performs the normal distribution approximation on both the target and source point clouds and finds the correspondences using exact nearest neighbor search with a kd-tree.
NDT tends to be more computationally efficient and GICP tends to be more accurate.

Feature-based methods extract geometric features, such as a line segment, plane, and point, from the input range data, and efficiently determine the correspondences.
LOAM\cite{LOAM-SLAM} was the first feature-based method to perform fast and accurate odometry calculations using LiDAR.
It significantly reduces the number of points required in the localization phase using feature matching.
LeGO-LOAM\cite{LeGO-LOAM-SLAM} further speeds up LOAM by relying only on good features to perform feature selection, and is one of the fastest methods currently available\cite{LiTAMIN}.

To achieve faster registration, some methods leverage GPU computation power, including SuMa\cite{SuMa-SLAM}, Elastic-Fusion\cite{ElasticFusion}, Elastic-LiDAR Fusion\cite{Elastic-SLAM}, and Droeschel et al.'s method\cite{ContSLAM}.
They approximate the shape of the range data as a set of small disks called a Surfel\cite{Surfel}.
A Surfel is a point-based rendering method\cite{PntRender}, which is designed to render 3D shapes with a point cloud instead of a polygon mesh, and is suitable for GPU processing.
It thus allows the performance of fast point-to-plane ICP\cite{Efficient-ICP} via the projective data association using hardware support.

Deep neural network-based approaches to LiDAR odometry\cite{LO-Net,DeepLO,DeepLO_Zero} have also come into fashion.
LO-Net\cite{LO-Net} is an end-to-end LiDAR-odometry method.
Although there is less variation in the data used for training and testing, it demonstrates accuracy similar to that of conventional approaches on a limited dataset.
The main computation in LO-Net is the convolutional tensor operation, which makes it easy to parallelize point-by-point processing, and is also scalable for the future evolution of GPUs.
However, whether end-to-end odometry estimation works on unlearned environments and motion has not been investigated sufficiently, and further research is needed.

Current LiDAR SLAM methods require roughly O(N) or O(Nlog(N)) for N points, and theoretically, different approaches should be introduced to improve the algorithm.
The claim in the present study is simple and straightforward: The number of points required for ICP-based methods should be small.
Generally, the accuracy of ICP-based methods degrades as the number of points is reduced; hence, an approach should be found to solve the trade-off problem between speed and accuracy.

\section{Method}

In this section, we describe the differences between the proposed method and the conventional method, LiTAMIN\cite{LiTAMIN}; the differences are the method used to reduce the number of points and the cost function used for the ICP.

\subsection{Reduction of the number of points}

As shown in Figure \ref{fig:overview}, LiTAMIN voted a group of input points into the voxel grids, aligned them using the means of the voting points, and integrated the point clouds into the voxel map.
The proposed method performs SLAM in a similar manner, but the difference is that it uses covariance, not just the mean, for the voting results of the input point groups.
Whereas LiTAMIN was a point-to-normal distribution mapping, the proposed method extends it to normal distribution-to-normal distribution mapping.
This is intended to improve accuracy by considering the spread of the distributions.
The proposed method increases each voxel size and reduces the number of points, which significantly reduces the computational cost.
Moreover, it avoids the loss of accuracy by considering the shape of the distribution instead of the points.

\subsection{ICP cost function applied with symmetric KL-divergence}

Table \ref{tbl:ICP} shows the cost functions of ICP for existing methods and the proposed method.
The difference between the proposed method and other methods is that the cost takes into account not only the distance between the points but also the shapes of the distributions.
Although other methods, such as NDT\cite{NDT-ICP}, GICP\cite{G-ICP}, and LiTAMIN\cite{LiTAMIN}, that take covariance into account have been proposed, in practice, they only evaluate the distance by weighting the inverse of the covariance.
The proposed method simultaneously evaluates the weighted distance in the first term and the difference in distribution shape in the second term.
For example, if the distances between points are small but the shape of the distribution does not match, the cost is designed to be large.
This cost was derived from the KL-divergence $D_{KL}(p\|q)$ of two Gaussian distributions $p$ and $q$ \cite{KLD,KLD2}:
{
\begin{equation}
\begin{split}
&D_{KL}(p \| q) = \int p(x) \log {p(x) \over q(x)} dx \propto \\
& (\mu_q-\mu_p)^T C_q^{-1} (\mu_q-\mu_p) + {\rm Tr} (C_q^{-1}C_p ) - d + \log {|C_q| \over |C_p|}.
\end{split}
\end{equation}
}
%
where $\mu_p$ and $\mu_q$ are means, $C_p$ and $C_q$ are covariance matrices, ${\rm Tr}(\cdot)$ is the matrix trace, and $d$ is the dimension of $x$.
KL-divergence is a measure of the difference between distributions, which represents not only the difference between the mean values but also the difference between the shapes of the distributions.
The aim of KL-divergence is to make a robust registration that takes into account the shape of the distribution.

KL-divergence is, however, not symmetric with $D_{KL}(p\|q) \neq D_{KL}(q\|p)$: it is generally considered not to be a distance.
Because ICP is a distance-minimizing algorithm and requires a more appropriate metric, $D_{SymKL}(p\|q)$ is used, which introduces the following symmetry in this study:
%
{
\begin{equation}
\begin{split}
D_{SymKL}(p \| q) \;=\;\; &  (\mu_q-\mu_p)^T (C_q+C_p)^{-1} (\mu_q-\mu_p) + \\
                          &  {\rm Tr} (C_q^{-1}C_p ) + {\rm Tr} (C_p^{-1}C_q ) - 2d.
%
%
\end{split}
\end{equation}
}
Our cost function is derived by applying Frobenius normalization to $D_{SymKL}(p\|q)$ and introducing a rigid body transformation $R$ and $t$.
To further address the outliers, the ICP error $E_{ICP}$ and distribution shape error $E_{Cov}$ are set as follows:
\begin{eqnarray}
E_{ICP} &=& \left(q- (R p + t)\right)^T C_{qp} \left(q- (R p + t)\right)\\
E_{Cov} &=& (\mathrm{Tr}(R C_{p}^{-1} R^T C_{q}) + \mathrm{Tr}(C_{q}^{-1} R C_{p} R^T) -6)^2, 
\end{eqnarray}
where
\begin{equation}
C_{qp} = \cfrac{(C_{q}+R C_{p} R^T + \lambda I)^{-1}}{\|(C_{q}+R C_{p} R^T + \lambda I)^{-1}\|_F}.  \nonumber
\end{equation}
Additionally, the weights are as follows:
{
\begin{eqnarray}
w_{ICP} &=& 1 - \frac{E_{ICP}}{E_{ICP}+\sigma_{ICP}^2},\\
w_{Cov} &=& 1 - \frac{E_{Cov}}{E_{Cov}+\sigma_{Cov}^2}.
\end{eqnarray}
}
Note that $\|\cdot\|_F$ is the Frobenius norm, $R$ and $t$ are estimates of the rigid body transformation, and $\sigma_{ICP} $ and $\sigma_{Cov}$ are acceptable error values.
$w_{ICP}$ and $w_{Cov}$ approach $1$ if the error is less than or equal to an acceptable value, and $0$ if it is greater than or equal to an acceptable value.

In Table \ref{tbl:ICP}, the first term of the proposed method can be regarded as the ICP cost considering the covariance of the two distributions at LiTAMIN.
The major difference from LiTAMIN is the second term that represents the difference in distribution shape.
In this study, this term is introduced so that accuracy does not decrease, even if the number of points is greatly reduced because of the large voxel size.
It is possible to calculate only the first term, that is, ICP cost.
Hence, the difference between the case of the first term alone, and the case of the first and second terms combined in experiments is investigated.

\subsection{Implementation and parameters}

In this study, the Newtonian method was used to optimize the cost function of the proposed method.
Because the second term of the cost function is not a squared error, it needs to be found up to the Hessian, and Newtonian optimization was used rather than the Gaussian--Newtonian method.
The damping factor of the Levenberg--Marquardt method\cite{SBA} was not used in this study because the calculations were stable without it.

The acceptable values of $\sigma_{ICP}$ and $\sigma_{Cov}$ were empirically set to $0.5$ and $3$, respectively.
$sigma_{ICP}$ corresponds to the Mahalanobis distance to $C_{qp}$, which means that correspondence points below $0.5$ are trusted.
$\sigma_{Cov}$ corresponds to $E_{Cov}$; if $C_{q}$ and $R C_{p} R^T$ are identical, $ E_{Cov}=\mathrm{Tr}(I)+\mathrm{Tr}(I)-6 $ should be $ 0$.
In $D_{SymKL}(p\|q)$, $-2d$ is a term to make the minimum value $0$ of $D_{SymKL}$.
$\sigma_{Cov}=3$ was set to allow for about half the error that would be allowed in the absence of this term.
The parameter $lambda$ of Frobenius normalization was set to $10^{-6}$, as in LiTAMIN. 

LiTAMIN used the occupancy probability\cite{ProbRobo} for weight $w$, whereas the proposed method used $w_{ICP}$ and $w_{Cov}$ instead.
For loop closure, the proposed ICP cost was used; the same parameters and implementations as in LiTAMIN were used for the other elements.
The corresponding points for ICP were searched using a kd-tree in addition to LiTAMIN.
Regarding the map representation, voxel maps were used in addition to LiTAMIN.
Voxel maps were used to reduce the number of normal distributions that make up the map.

LiTAMIN implemented tracking and mapping in separate threads, but the proposed method combined them in a single thread.
This is because it had sufficient computational efficiency, and to reduce the overhead of communication between threads.
Loop closure and the graph optimizer were implemented similarly to LiTAMIN.

\section{Experiments}

\begin{figure*}[!t]
\centering
\vspace{2mm}
\includegraphics[width=0.925\linewidth]{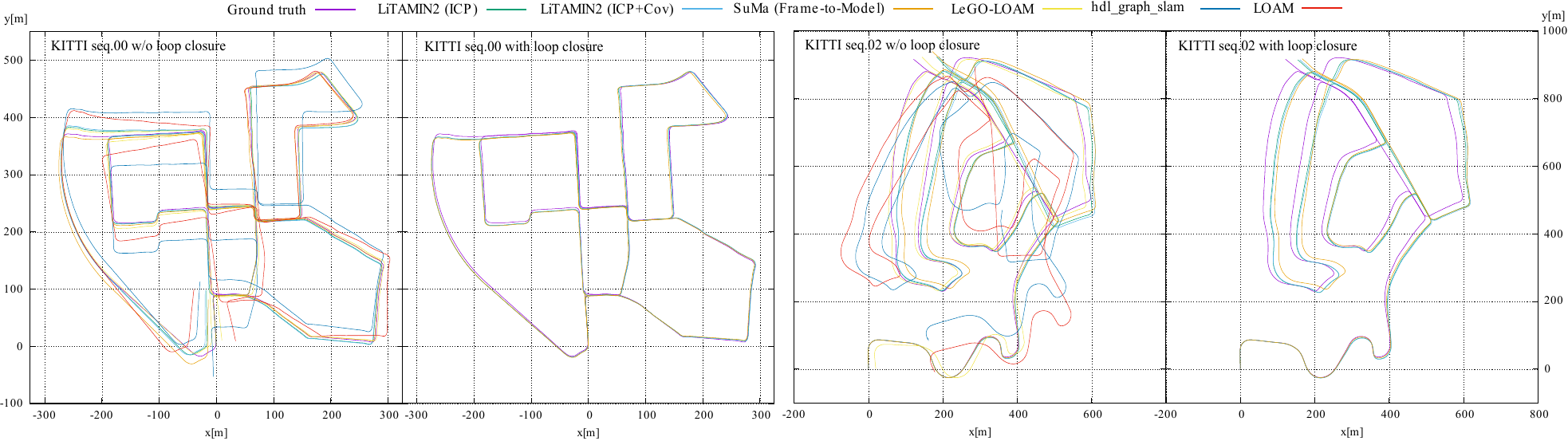}
\caption{Comparison of trajectories between GT data and each method. }
\label{fig:00-loop}
\vspace{-3mm}
\end{figure*}

\subsection{Comparison}

Several state-of-the-art methods were selected as the competitors that used different speeding up algorithms, specifically, LiTAMIN, SuMa, LeGO-LOAM, LOAM, hdl-graph-slam\cite{hdl-graph-SLAM}, LO-Net and DeepLO.
A detailed evaluation is provided on the SuMa project page, thus this was referred to in the study, whereas the computation time was acquired by the researchers running the open source.
The open sources of LeGO-LOAM, LOAM, and hdl-graph-slam were also used to obtain the trajectories and to measure the computation speeds in the following experiments.
Regarding LO-Net, the results in the original paper were used.

Each experiment was conducted using a desktop PC with an Intel Core i9-9900K with 32 GB RAM and an NVIDIA GeForce RTX 2080 Ti.

\subsection{Evaluation benchmark and criteria}

The KITTI Vision Benchmark was used in the experiments, which contains point clouds captured by an on-board Velodyne HDL-64E S2 in several environments.
It thus allows the evaluation of the trajectories obtained by any SLAM methods.
The provided point clouds were already deskewed, thus they were fed directly into the proposed method and the competitors.

The performance of each method was evaluated based on the following three criteria:
\begin{enumerate}
	\item {\bf KITTI stats: }
        The KITTI Vision Benchmark\cite{KITTI} statistics, that is, KITTI stats, were used in the accuracy evaluation.
	These criteria enable the evaluation of the quality of the estimated trajectory using the relative relations against the ground truth.
	In this study, the translation and rotation errors were calculated in that manner for different lengths, specifically, every 100 m up to 800 m, and the average of the errors was computed.
	Code provided by the benchmark was used to calculate the KITTI stats.

	\item {\bf Absolute Trajectory Error (ATE):}
	To evaluate the loop closing performance of each method, the ATE\cite{ErrMetric} was also calculated.
	ATE is an indicator of the absolute position and attitude error against the ground truth.
	The KITTI stats are calculated as an average of errors in sub-trajectories, which may underrate the effect of loop closing; however, the ATE allows a comparison of the entire shape of trajectories modified by loop closing based on the absolute error evaluation.

        \item {\bf Total time and frame rate:}
        As an index of the computational efficiency, the total time taken to process all sequences of the KITTI Vision Benchmark including the loop closing was calculated.
        The frame rate of the odometry was also presented to evaluate the speed of the position estimation, which may be important for some real-time applications.
\end{enumerate}

\subsection{Ablation study}

The proposed method approximates the sub-point cloud voted in each voxel to a normal distribution.
Because the voxel size significantly affects performance, the proposed method was thoroughly evaluated for different voxel sizes, as shown in Table \ref{tbl:LiTAMIN2}.
Additionally, Table \ref{tbl:LiTAMIN2} shows the average reduction percentage from the original scan points using voxelization.

From the KITTI stats, it is not always the case that the finer the voxel, the better the accuracy.
Note that the voxel size was fixed at 3 meters in the following experiments because the best performance was achieved with this value.


\subsection{Comparison study}

Table \ref{tbl:KITTI-Stats} shows a comparison of KITTI stats. 
For SuMa, the trajectory datasets, frame-to-frame, frame-to-model, and frame-to-model with loop closure, obtained from the authors' project page were compared.
For LeGO-LOAM and hdl\_graph\_slam, loop closure was implemented, but because loop detection did not occur in our experiments, the results were not listed in Table \ref{tbl:KITTI-Stats}.
For LOAM, the results of measurements using open sources in their paper and the results of the original paper were considered. The statistics of the individual sequences were listed in the original paper, but the final resultant KITTI stats were not listed in Table \ref{tbl:KITTI-Stats} because they were not listed in the original paper.
LO-Net was not listed in Table \ref{tbl:KITTI-Stats} because statistics for individual sequences were provided in \cite{LO-Net}, but the average of all final error values was not provided.

Table \ref{tbl:ATE} represents a comparison of ATEs.
The results for SuMa were evaluated for trajectories taken from the authors' project page, in addition to Table \ref{tbl:LiTAMIN2}.
For LOAM, the results for the open source software are shown.
LO-Net was excluded in Table \ref{tbl:ATE} because the results of ATE were not included in the original paper.

Table \ref{tbl:SpeedResult} shows the processing time taken to create the map using each sequence of KITTI and the average frame rate of the odometry process.
The proposed method, LiTAMIN, and SuMa show the processing time including loop closing because loop closing in these methods was successful for all sequences.
The SuMa results were obtained on the computer used in this study using open source because the computer specification in SuMa's original paper was different.

Figure \ref{fig:00-loop} shows the comparison of trajectories for each method. 
The second and forth figures from the left show the loop closure results for the proposed method and SuMa, but not other methods that could not detect loops.


{
\begin{table*}[t]\renewcommand{\arraystretch}{1.2}
\begin{center}
\vspace{2mm}
\caption{Performance table for LiTAMIN2 with loop closure.}
\label{tbl:LiTAMIN2}
\setlength{\tabcolsep}{0.75mm}
{\scriptsize
\begin{tabular}{lc|c|cccccccccccccccccccc}
Voxel size &[m] & ICP cost & 0.5 & 1.0 & 1.5 & 2.0 & 2.5 & 3.0 & 3.5 & 4.0 & 4.5 & 5.0 & 5.5 & 6.0 & 6.5 & 7.0 & 7.5 & 8.0 & 8.5 & 9.0 & 9.5 & 10 \\ \midrule[1pt]
Total time for all seq.  &[sec]      &         & 3063 & 566 & 222 & 121 & 80 & 58 & 48 & 41 & 41 & 39 & 39 & 43 & 37 & 38 & 33 & 37 & 34 & 37 & 35 & 35\\
Odometry frame rate      &[FPS]      & LiTMIN2 & 7.9 & 45 & 118 & 225 & 363 & 510 & 664 & 814 & 992 & 1122 & 1255 & 1364 & 1387 & 1432 & 1420 & 1355 & 1302 & 1251 & 1122 & 1074\\ 
KITTI stats: rotation    &[deg/100m] & ICP     & 1.51 & 0.42 & 0.35 & 0.37 & 0.38 & 0.38 & 0.40 & 0.48 & 0.54 & 0.61 & 0.73 & 0.77 & 0.90 & 1.08 & 1.11 & 1.34 & 1.58 & 1.60 & 1.80 & 2.16\\
KITTI stats: translation &[\%]       &         & 6.01 & 1.42 & 1.11 & 1.00 & 0.92 & 0.89 & 0.88 & 0.99 & 1.09 & 1.36 & 1.57 & 1.67 & 1.88 & 2.32 & 2.49 & 3.26 & 4.76 & 4.37 & 5.52 & 5.77\\ \hline
Total time for all seq.  &[sec]      &         &4656 & 1058 & 453 & 251 & 161 & 119 & 94 & 81 & 70 & 65 & 63 & 64 & 69 & 64 & 66 & 68 & 68 & 71 & 80 & 88\\
Odometry frame rate      &[FPS]      & LiTMIN2 & 5.3 & 25 & 60 & 110 & 173 & 239 & 308 & 369 & 428 & 470 & 490 & 500 & 498 & 504 & 468 & 449 & 400 & 375 & 327 & 301 \\ 
KITTI stats: rotation    &[deg/100m] & ICP+Cov & 1.46 & 0.56 & 0.38 & 0.36 & 0.37 & 0.33 & 0.36 & 0.44 & 0.50 & 0.58 & 0.77 & 0.85 & 0.97 & 1.35 & 1.36 & 1.71 & 1.89 & 1.94 & 2.41 & 2.44\\
KITTI stats: translation &[\%]       &         & 5.85 & 1.84 & 1.24 & 0.96 & 0.98 & 0.85 & 0.91 & 0.97 & 1.11 & 1.34 & 1.58 & 1.75 & 2.06 & 2.79 & 2.87 & 3.74 & 4.87 & 4.84 & 6.99 & 7.02\\ \hline
Pointcloud reduction ratio &[\%]&              & 5.67 & 2.35 & 1.37 & 0.89 & 0.64 & 0.50 & 0.41 & 0.34 & 0.29 & 0.25 & 0.23 & 0.20 & 0.18 & 0.16 & 0.15 & 0.14 & 0.13 & 0.12 & 0.11 & 0.10 \\
\bottomrule
\end{tabular}
}
\end{center}
{\tiny
{\bf Total time for all seq.} is the total processing time against all frames of all sequences. 
{\bf Odometry frame rate} is the average frame rate for odometry processing for all frames. 
{\bf KITTI stats} is the evaluation value of the KITTI Vision Benchmark against each voxel size. 
{\bf Point cloud reduction ratio} is the ratio from the number of points from original raw scans.  
}
\end{table*}
}

{
\begin{table*}[!ht]\renewcommand{\arraystretch}{1.2}
\begin{center}
\vspace{1mm}
\caption{KITTI stats for each SLAM method.}
\label{tbl:KITTI-Stats}
\setlength{\tabcolsep}{0.85mm}
{\scriptsize
\begin{tabular}{c|ccccccccccccc}
Method           & Loop    & Seq. 00 & Seq. 01 & Seq. 02 & Seq. 03 & Seq. 04 & Seq. 05 & Seq. 06 & Seq. 07 & Seq. 08 & Seq. 09 & Seq. 10 & KITTI stats\\
(Num. of frames) & closure & (4541)  & (1101)  & (4661)  & (801)   & (271)   & (2761)  & (1101)  & (1101)  & (4071)  & (1591)  & (1201)  & [deg/100m] / [\%] \\ \midrule[1pt]
LiTAMIN2 (ICP+Cov)      &\Mns & 0.36/0.78 & 0.46/2.10 & 0.37/0.95 & 0.48/0.96 & 0.52/1.05 & 0.31/0.55 & 0.33/0.55 & 0.49/0.48 & 0.35/1.01 & 0.40/0.69 & 0.47/0.80 & 0.38 / 0.88\\ 
LiTAMIN2 (ICP+Cov)      &\Chk & 0.28/0.70 & 0.46/2.10 & 0.32/0.98 & 0.48/0.96 & 0.52/1.05 & 0.25/0.45 & 0.34/0.59 & 0.32/0.44 & 0.29/0.95 & 0.40/0.69 & 0.47/0.80 & 0.33 / 0.85\\ \hline
LiTAMIN2 (ICP)          &\Mns & 0.42/0.75 & 0.40/1.88 & 0.45/0.99 & 0.43/0.84 & 0.41/0.90 & 0.32/0.74 & 0.23/0.45 & 0.57/0.55 & 0.55/1.25 & 0.32/0.74 & 0.59/1.36 & 0.45 / 0.95\\ 
LiTAMIN2 (ICP)          &\Chk & 0.33/0.70 & 0.40/1.88 & 0.37/0.92 & 0.43/0.84 & 0.41/0.90 & 0.28/0.50 & 0.31/0.50 & 0.34/0.43 & 0.48/1.16 & 0.27/0.81 & 0.59/1.36 & 0.38 / 0.89\\ \hline
LiTAMIN                 &\Mns & 0.46/0.91 & 0.45/11.3 & 0.46/1.30 & 0.56/1.17 & 0.47/18.7 & 0.39/0.75 & 0.29/0.59 & 0.34/0.48 & 0.42/1.04 & 0.45/0.99 & 0.90/3.78 & 0.46 / 1.60\\ 
LiTAMIN                 &\Chk & 0.41/0.95 & 0.45/11.3 & 0.45/1.25 & 0.56/1.17 & 0.47/18.7 & 0.35/0.70 & 0.32/0.63 & 0.33/0.45 & 0.37/1.03 & 0.43/1.06 & 0.90/3.78 & 0.43 / 1.59 \\ \hline
SuMa (Frame-to-Frame)   &\Mns & 0.92/2.11 & 1.21/4.31 & 0.78/2.32 & 0.73/1.63 & 1.05/11.9 & 0.76/1.46 & 0.57/0.89 & 1.09/1.87 & 0.95/2.56 & 0.76/1.99 & 0.94/2.15 & 0.88 / 2.19\\  
SuMa (Frame-to-Model)   &\Mns & 0.30/0.72 & 0.47/1.77 & 0.39/1.06 & 0.46/0.57 & 0.27/0.39 & 0.23/0.50 & 0.14/0.39 & 0.33/0.37 & 0.35/1.01 & 0.25/0.47 & 0.27/0.69 & 0.33 / 0.84\\  
SuMa (Frame-to-Model)   &\Chk & 0.22/0.64 & 0.47/1.77 & 0.41/1.23 & 0.46/0.57 & 0.27/0.39 & 0.20/0.42 & 0.28/0.51 & 0.52/0.65 & 0.35/1.15 & 0.20/0.57 & 0.27/0.69 & 0.32 / 0.89\\ \hline 
LeGO-LOAM               &\Mns & 1.05/2.17 & 1.02/13.4 & 1.01/2.17 & 1.18/2.34 & 1.01/1.27 & 0.74/1.28 & 0.63/1.06 & 0.81/1.12 & 0.94/1.99 & 0.98/1.97 & 0.92/2.21 & 1.00 / 2.49\\ \hline
hdl\_graph\_slam        &\Mns & 1.00/3.92 & 7.62/93.5 & 1.84/11.2 & 0.92/1.71 & 1.21/96.0 & 0.69/1.41 & 0.94/11.1 & 1.10/1.28 & 0.99/2.17 & 0.93/4.32 & 0.75/2.36 & 1.49 / 9.57\\ \hline
LOAM                    &\Mns & 0.91/1.92 & 0.71/2.69 & 1.49/4.05 & 0.63/1.38 & 0.66/1.21 & 0.59/1.17 & 0.36/0.82 & 0.62/0.91 & 0.62/1.43 & 0.57/1.21 & 0.61/1.53 & 0.90 / 2.13 \\ 
LOAM (from \cite{LOAM-SLAM-Paper}) &\Mns &   - /0.78 &   - /1.43 &   - /0.92 &   - /0.86 &   - /0.71 &   - /0.57 &   - /0.65 &   - /0.63 &   - /1.12 &   - /0.77 &   - /0.79 &   - / - \\ \hline
LO-Net (Frame-to-Frame) &\Mns & 0.72/1.47 & 0.47/1.36 & 0.71/1.52 & 0.66/1.03 & 0.65/0.51 & 0.69/1.04 & 0.50/0.71 & 0.89/1.70 & 0.77/2.12 & 0.58/1.37 & 0.93/1.80 & - / - \\
LO-Net (Frame-to-Model) &\Mns & 0.42/0.78 & 0.40/1.42 & 0.45/1.01 & 0.59/0.73 & 0.54/0.56 & 0.35/0.62 & 0.33/0.55 & 0.45/0.56 & 0.43/1.08 & 0.38/0.77 & 0.41/0.92 & - / - \\ \hline
DeepLO (from \cite{DeepLO}) &\Mns & - / - & - / - & - / - & - / - & - / - & - / - & - / - & - / - & - / - & 1.95/4.87 & 1.83/5.02 & - / - \\
\bottomrule
\end{tabular}
}
\end{center}
{\tiny
For LiTAMIN2, the size of the voxel was 3 m from the best accuracy result in Table \ref{tbl:LiTAMIN2}. 
The marks \ChkMini and {\MnsMini} represent with (\ChkMini) and without (\MnsMini) loop closure, respectively, for each method.
}
\end{table*}
}

{
\begin{table*}[!ht]\renewcommand{\arraystretch}{1.2}
\begin{center}
\caption{Absolute trajectory error for each SLAM method.}
\label{tbl:ATE}
\setlength{\tabcolsep}{1.1mm}
{\scriptsize
\begin{tabular}{c|ccccccccccccc}
Method           & Loop    & Seq. 00 & Seq. 01 & Seq. 02 & Seq. 03 & Seq. 04 & Seq. 05 & Seq. 06 & Seq. 07 & Seq. 08 & Seq. 09 & Seq. 10 & Avg. of all frames\\
(Num. of frames) & closure & (4541)  & (1101)  & (4661)  & (801)   & (271)   & (2761)  & (1101)  & (1101)  & (4071)  & (1591)  & (1201)  & [deg] / [m] \\ \midrule[1pt]
LiTAMIN2 (ICP+Cov)    &\Mns & 1.6/5.8 & 3.5/15.9 & 2.7/10.7 & 2.6/0.8 & 2.3/0.7 & 1.1/2.4 & 1.1/0.9 & 1.0/0.6 & 1.3/2.5 & 1.7/2.1 & 1.2/1.0 & 1.8 / 5.1\\ 
LiTAMIN2 (ICP+Cov)    &\Chk & 0.8/1.3 & 3.5/15.9 & 1.3/3.2 & 2.6/0.8 & 2.3/0.7 & 0.7/0.6 & 0.8/0.8 & 0.6/0.5 & 0.9/2.1 & 1.7/2.1 & 1.2/1.0 & 1.2 / 2.4\\ \hline
LiTAMIN2 (ICP)        &\Mns & 1.8/5.4 & 3.1/13.8 & 3.4/12.1 & 2.4/0.7 & 1.9/0.6 & 1.4/3.6 & 0.7/0.7 & 1.3/0.9 & 3.0/5.9 & 1.9/2.8 & 1.5/1.8 & 2.3 / 6.0\\ 
LiTAMIN2 (ICP)        &\Chk & 0.8/1.2 & 3.1/13.8 & 1.3/3.0 & 2.4/0.7 & 1.9/0.6 & 0.7/0.7 & 0.8/0.6 & 0.6/0.4 & 2.2/4.5 & 0.8/1.3 & 1.5/1.8 & 1.3 / 2.6\\ \hline
LiTAMIN               &\Mns & 2.0/4.7 & 3.0/84.3 & 2.4/9.7 & 3.4/0.8 & 1.4/21.3 & 1.4/2.3 & 0.7/0.9 & 0.7/0.5 & 1.9/3.5 & 1.4/1.6 & 1.7/1.7 & 1.9 / 8.3\\ 
LiTAMIN               &\Chk & 1.1/1.5 & 3.0/84.3 & 1.8/3.7 & 3.4/0.8 & 1.4/21.3 & 0.9/1.0 & 0.8/0.8 & 0.6/0.3 & 1.6/2.8 & 1.3/1.4 & 1.7/1.7 & 1.5 / 6.2\\ \hline
SuMa (Frame-to-Frame) &\Mns & 6.4/19.7 & 8.2/34.9 & 5.4/21.3 & 4.1/1.2 & 3.4/13.4 & 2.9/5.1 & 1.5/2.0 & 2.1/2.9 & 6.2/15.9 & 2.4/5.0 & 2.4/3.4 & 4.8 / 14.1\\  
SuMa (Frame-to-Model) &\Mns & 1.0/2.9 & 3.2/13.8 & 2.2/8.4 & 1.5/0.9 & 1.8/0.4 & 0.7/1.2 & 0.4/0.4 & 0.7/0.5 & 1.5/2.8 & 1.1/2.9 & 0.8/1.3 & 1.4 / 3.9\\  
SuMa (Frame-to-Model) &\Chk & 0.7/1.0 & 3.2/13.8 & 1.7/7.1 & 1.5/0.9 & 1.8/0.4 & 0.5/0.6 & 0.7/0.6 & 1.1/1.0 & 1.2/3.4 & 0.8/1.1 & 0.8/1.3 & 1.1 / 3.2\\ \hline 
LeGO-LOAM             &\Mns & 2.8/6.3 & 3.8/119.4 & 4.1/14.7 & 4.1/0.9 & 3.3/0.8 & 1.9/2.8 & 1.4/0.8 & 1.5/0.7 & 2.5/3.5 & 2.2/2.1 & 1.9/1.8 & 2.8 / 11.1\\ \hline
hdl\_graph\_slam      &\Mns & 5.4/41.8 & 34.0/635.8 & 22.3/153.0 & 2.3/1.0 & 3.4/93.4 & 2.5/5.7 & 3.3/43.0 & 2.2/1.6 & 6.2/13.8 & 4.6/15.9 & 1.8/3.5 & 9.3 / 76.7\\ \hline
LOAM                  &\Mns & 5.8/19.4 & 6.1/21.0 & 21.7/111.6 & 3.3/1.0 & 2.2/0.5 & 2.2/4.6 & 0.9/1.1 & 1.2/1.3 & 3.0/6.7 & 1.9/5.3 & 1.5/1.9 & 7.0 / 29.7\\
\bottomrule
\end{tabular}
}
\end{center}
{\tiny
For LiTAMIN2, the size of the voxel was 3 m from the best accuracy result in Table \ref{tbl:LiTAMIN2}. 
The marks \ChkMini and {\MnsMini} represent with (\ChkMini) and without (\MnsMini) loop closure, respectively, for each method.
}
\end{table*}
}

{
\begin{table*}[!ht]\renewcommand{\arraystretch}{1.2}
\begin{center}
\caption{Computation time for building a map, and the odometry frame rate.}
\label{tbl:SpeedResult}
\setlength{\tabcolsep}{0.7mm}
{\scriptsize
\begin{tabular}{c|ccccccccccccc}
Method           & Loop    & Seq. 00 & Seq. 01 & Seq. 02 & Seq. 03 & Seq. 04 & Seq. 05 & Seq. 06 & Seq. 07 & Seq. 08 & Seq. 09 & Seq. 10 & Total time / Avg. rate\\
(Num. of frames) & closure & (4541)  & (1101)  & (4661)  & (801)    & (271)   & (2761)  & (1101)  & (1101)  & (4071)  & (1591)  & (1201)  & [sec] / [FPS] \\ \midrule[1pt]
LiTAMIN2 (ICP)        &\Chk &  8.4/597  &  6.1/189  &  13.3/545  &  2.1/434  &  1.2/282  &  4.9/610  &  3.4/370  &  2.2/625  &  9.9/431  &  4.2/432  &  2.0/599  &   58 / 508.9\\ 
LiTAMIN2 (ICP+Cov)    &\Chk & 15.9/299  & 15.9/70.1 &  24.2/243  &  4.4/193  &  2.6/109  &  9.4/305  &  7.3/159  &  3.8/323  & 21.7/191  &  9.4/181  &  4.4/294  &  119 / 238.8\\ 
LiTAMIN               &\Chk & 87.9/53.0 & 70.4/15.8 & 108.1/44.0 & 18.9/43.4 & 10.1/27.3 & 53.3/53.0 & 32.7/34.4 & 20.2/56.2 & 97.4/42.9 & 43.3/37.5 & 23.5/52.5 &  566 / 45.2\\ 
SuMa (Frame-to-Model) &\Chk & 90.1/55.1 & 20.2/57.1 &  77.6/65.7 & 12.5/65.6 &  5.4/51.9 & 57.2/54.6 & 21.4/55.4 & 18.7/65.7 & 74.9/58.0 & 31.4/52.5 & 22.4/55.2 &  432 / 58.4\\ 
LeGO-LOAM             &\Mns & 78.4/69.9 & 27.5/69.4 &  91.8/64.5 & 16.5/62.8 &  5.4/64.9 & 50.1/67.7 & 24.7/64.9 & 17.5/73.4 & 80.1/65.9 & 32.7/63.2 & 21.1/66.8 &  445 / 66.1\\ 
hdl\_graph\_slam      &\Mns &  800/5.7  &  197/5.6  &  1107/4.2  &  163/4.9  & 53.1/5.1  &  596/4.6  &  293/3.8  &  209/5.3  &  819/5.0  &  440/3.6  &  252/4.8  & 4929 / 4.8\\ 
LOAM                  &\Mns & 414/11.0  & 83.2/13.5 &   448/10.4 & 75.2/10.4 & 24.6/10.2 &  266/10.3 &  102/10.6 & 97.6/11.1 &  383/10.6 &  151/10.4 &  114/10.4 & 2156 / 10.7\\
\bottomrule
\end{tabular}
}
\end{center}
{\tiny
For LiTAMIN2, the size of the voxel was 3 m from the best accuracy result in Table \ref{tbl:LiTAMIN2}. 
The marks \ChkMini and {\MnsMini} represent with (\ChkMini) and without (\MnsMini) loop closure, respectively, for each method.
}
\vspace{-1mm}
\end{table*}
}

\section{Discussion}

From the results in Table \ref{tbl:LiTAMIN2}, the proposed method was most accurate when the voxel size was 3 m and ICP+Cov was used as the cost function.
The frame rate of odometry at that time was 510 fps for the ICP cost alone and 239 fps for the ICP+Cov cost, which was much faster than the conventional methods in Table \ref{tbl:SpeedResult}.
This could simply be because the number of points used for registration was greatly reduced by voxel voting.
Moreover, the proposed method was as accurate as the most accurate method, SuMa, as can be seen from Table \ref{tbl:KITTI-Stats}, despite the significant decrease in the number of points.
We believe that this is the result of the original design intention, and the result of taking the shape of the distribution into account using symmetric KL-divergence.

According to the odometry frame rates in Table \ref{tbl:LiTAMIN2}, a comparison of the case with ICP costs alone and the case with ICP+Cov costs confirms that the accuracy of the rotation using Cov costs was slightly higher than the results using ICP alone.
The reason could be that the shape of the normal distribution was more rotationally constrained.
${\rm Tr} (C_q^{-1}C_p )$ in Table \ref{tbl:ICP} is a cost whose value decreased as the main axes of the normal distribution coincided, and we believe that this is also the result of our intention.
However, as shown in the results of the odometry frame rates, the processing time was about twice as long in terms of calculation cost, so the ICP cost alone is sufficient if accuracy is not important.
The choice of which cost to use depends on the application.

If the same accuracy as that of LiTAMIN by the conventional method is sufficient, then Table \ref{tbl:LiTAMIN2} confirms that the proposed method could achieve the same level of accuracy as LiTAMIN, even for voxels with a roughness of about 6 m.
The frame rate of the proposed method exceeded 1000 fps for the ICP cost only, which is the fastest ever achieved.
The results are applicable to the scale of the KITTI Vision Benchmark environment, which is similar to driving on a roadway outdoors.
In indoor and more confined environments, the size of the voxels should be chosen appropriately.

Table \ref{tbl:ATE} is the result of evaluating the consistency of the entire trajectory.
The proposed method achieved better accuracy than SuMa after loop closing, and the effect of the loop closing correction was significant.
We can confirm that the proposed method had a relatively large decrease in the error in the results before and after loop closing for SuMa.
This is because the loop closing method of LiTAMIN, the predecessor of the proposed method, worked properly, and the proposed ICP cost used for the detection of the loop constraint worked properly.
This can be confirmed by the large error reduction in the average of all frames of the proposed method in comparison with LiTAMIN.

\section{Conclusion}

In this study, an ICP method using symmetric KL-divergence was proposed to improve the speed of LiDAR-based SLAM significantly, and  it was compared with other state-of-the-art SLAM methods.
The proposed method achieved a computational speed of 500 fps to 1000 fps in the odometry frame rate, with the same level of accuracy as the other methods, which confirms that the proposed method is a great step forward from conventional methods.
This is because the number of points used for registration was significantly reduced by voting the point cloud from LiDAR into each voxel grid and approximating the voted sub-point cloud into a single normal distribution.
Although the proposed method reduced the number of points significantly, the ICP cost of the proposed symmetric KL-divergence allowed the data to be processed without reducing accuracy.
These results were based on the KITTI Vision Benchmark dataset, and we believe that we need to investigate how to determine the appropriate voxel size when the usage environment changes.


{\small
\bibliographystyle{IEEEtran}
\bibliography{egbib}
}



\end{document}